\documentclass[runningheads]{llncs}

\usepackage{eccv}

\usepackage{eccvabbrv}

\usepackage{graphicx}
\usepackage{float} 
\usepackage{booktabs}
\usepackage{xcolor}
\usepackage{pifont}

\newcommand{\cmark}{\textcolor{green!60!black}{\ding{51}}} 
\newcommand{\xmark}{\textcolor{red!70!black}{\ding{55}}}   
\usepackage{makecell}
\usepackage[accsupp]{axessibility}  

\usepackage{multirow} 
\usepackage{kotex}
\usepackage{enumitem}

\usepackage{hyperref}

\usepackage{orcidlink}

\usepackage{comment}
\usepackage{wrapfig}

\begin{document}

\title{
\textsc{JointHOI}: Jointly Generating Contact Maps Enhances Hand--Object Interaction Generation} 

\titlerunning{JointHOI}


\author{
Mingyeong Song\inst{1} \and
Jungbin Cho\inst{2,3} \and
Jisoo Kim\inst{2} \and
Ananya Bal\inst{3} \and
Kartik Sharma\inst{3} \and
Youngjae Yu\inst{4} \and
Laszlo A. Jeni\inst{3} \and
Junhyug Noh\inst{1}\thanks{Corresponding author: J. Noh (junhyug@ewha.ac.kr).}
}

\authorrunning{Song et al.}

\institute{
Ewha Womans University, Seoul, Korea \and
Yonsei University, Seoul, Korea \and
Carnegie Mellon University, Pittsburgh PA, USA \and
Seoul National University, Seoul, Korea
}

\maketitle

{\centering
\small
\textbf{Project Page:} \url{https://smk11602.github.io/JointHOI/}
\par}

\begin{figure}[H]
  \centering
  \includegraphics[width=\linewidth]{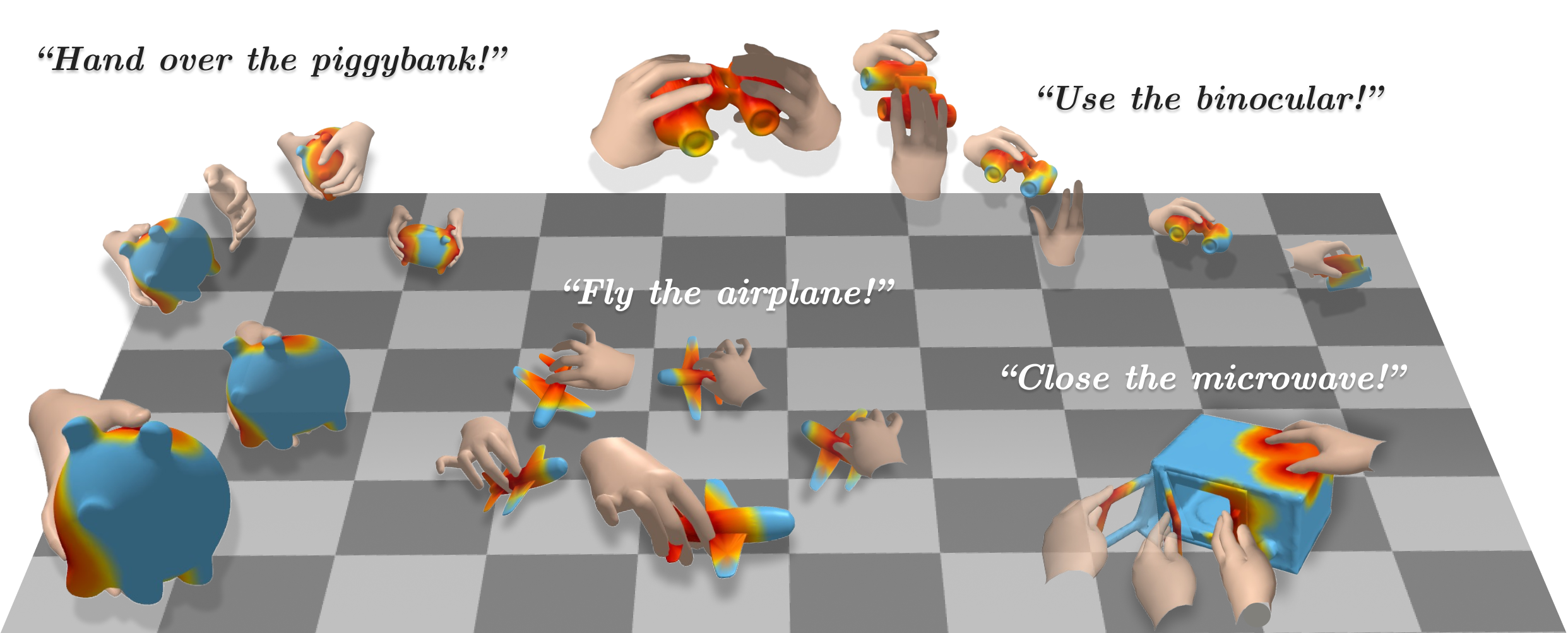}
\caption{\textsc{JointHOI} generates bimanual hand--object interaction sequences from text prompts, along with dynamic contact maps that capture time-varying hand--object proximity and help reduce artifacts such as floating and interpenetration.}
  \label{fig:teaser}
\end{figure}

\begin{abstract}
Text-driven hand--object interaction (HOI) generation is gaining attention for immersive applications and robotics, yet producing \emph{physically plausible} interactions remains challenging. Even when individual motions appear natural, small contact errors can cause conspicuous artifacts such as floating and interpenetration. Prior methods mitigate these issues using explicit contact cues or implicit grasp priors, but typically rely on multi-stage pipelines and fail to model \emph{temporally evolving} contact. We present \textsc{JointHOI}, a single-stage diffusion framework that jointly generates 3D hand--object motion and dynamic, distance-based contact maps from text. By treating contact as an auxiliary \emph{inner modality}, joint generation enables the model to learn contact--motion coupling during training. At inference, contact-guided sampling enforces consistency between generated contact maps and motion-implied geometry, improving temporal stability and reducing penetration and floating. Experiments on GRAB and ARCTIC demonstrate consistent improvements in text adherence and physical plausibility over prior methods.
\end{abstract}

\section{Introduction}
\label{sec:intro}

Hand--object interactions are defined by contact: where the hand touches the object, when contact forms or breaks, and how that contact evolves as the motion unfolds.
This makes contact fidelity central to HOI generation for robotics and AR/VR -- robots require precise, stable contacts for reliable manipulation, while AR/VR demands visually convincing interactions that withstand close human scrutiny.
Generating HOI from text is therefore not merely a motion synthesis problem; it requires producing \emph{contact-consistent} kinematics with high geometric fidelity.
Otherwise, even small errors manifest as interpenetration, floating fingers, or unstable grasps, artifacts that humans are sensitive to~\cite{burns2006hand,prachyabrued2012visual}.

Recent text-conditioned methods such as Text2HOI~\cite{cha2024text2hoi} take an important step toward semantic control, but realistic HOI remains challenging because interaction dynamics are still captured only indirectly.
The core difficulty is that contact in HOI is \emph{spatiotemporally structured}: it varies across the object surface and evolves continuously over time as hands approach, grasp, manipulate, and release.
Representations that collapse this structure -- either by removing temporal variation or by hiding geometry in a latent code -- make it difficult to maintain physical plausibility throughout a generated sequence.

Accordingly, recent works~\cite{cha2024text2hoi,Muchen_LatentHOI,christen2024diffh2o,hoigpt} augment motion generation with additional interaction modeling.
A common approach is to pretrain an autoencoder that implicitly captures hand--object relationships in a latent space.
While effective, this introduces a multi-stage training pipeline and lacks an explicit, physically grounded contact signal that can be inspected or enforced.
Alternatively, Text2HOI explicitly predicts contact maps and uses them as a conditioning signal for motion generation; however, its contact representation is \emph{static} and \emph{binary}, which cannot capture time-varying contact evolution and provides only a coarse notion of proximity.
Moreover, it relies on a multi-stage pipeline and an additional refinement step, further increasing complexity.
We summarize these design choices across recent text-driven HOI methods in \cref{tab:component_comparison}.

In this paper, we propose \textsc{JointHOI}, a single-stage joint diffusion framework that concurrently generates HOI motion and \emph{dynamic, distance-based} contact maps directly from text.
Our key idea is to treat contact as an \emph{inner modality} of hand--object motion: we co-generate 3D trajectories together with contact evolution within a unified diffusion process so the model can learn motion--contact dependencies explicitly rather than inferring them only implicitly from kinematics.
Concretely, we parameterize contact as a distance-to-surface field over a fixed set of object-surface anchors, yielding a dense, continuous signal that tracks contact evolution frame-by-frame.
Jointly modeling motion and contact mitigates error propagation in sequential pipelines and lets the model internalize physically grounded co-variation patterns (\eg, approach, slide, and release) directly through transformer self-attention.


\begin{table}[t!]
\centering
\footnotesize
\setlength{\tabcolsep}{3.8pt}
\renewcommand{\arraystretch}{1.15}

\setlength{\abovecaptionskip}{5pt}

\resizebox{\columnwidth}{!}{%
\begin{tabular}{l c c c c c c c}
\toprule
\textbf{Method} &
\makecell{\textbf{No}\\\textbf{post-ref.}$^{\dagger}$} &
\makecell{\textbf{Single-}\\\textbf{stage}$^{\ddagger}$} &
\makecell{\textbf{Implicit}\\\textbf{interaction}$^{\S}$} &
\makecell{\textbf{Explicit}\\\textbf{contact}$^{\P}$} &
\makecell{\textbf{Dynamic}\\\textbf{contact}} &
\makecell{\textbf{Distance}\\\textbf{contact}} &
\makecell{\textbf{Inner}\\\textbf{guidance}} \\
\midrule
Text2HOI~\cite{cha2024text2hoi}        & \xmark & \xmark & \xmark & \cmark & \xmark & \xmark & \xmark \\
DiffH2O~\cite{christen2024diffh2o}     & \cmark & \xmark & \xmark & \xmark & \xmark & \xmark & \xmark \\
LatentHOI~\cite{Muchen_LatentHOI}      & \cmark & \xmark & \cmark & \xmark & \xmark & \xmark & \xmark \\
HOIGPT~\cite{hoigpt}                   & \cmark & \xmark & \cmark & \xmark & \xmark & \xmark & \xmark \\
\midrule
\textbf{\textsc{JointHOI} (Ours)}               & \cmark & \cmark & \xmark & \cmark & \cmark & \cmark & \cmark \\
\bottomrule
\end{tabular}%
}
\caption{\textbf{Design matrix for text-driven HOI generation.}
\textsc{JointHOI} combines (i) \emph{single-stage} generation without post-refinement and
(ii) \emph{explicit} interaction modeling via co-generated \emph{dynamic, distance-based} contact maps,
which further enables inference-time \emph{inner guidance} for physically plausible hand--object motion.
$^{\dagger}$No post-refinement: no extra optimization/post-processing after generation.
$^{\ddagger}$Single-stage: no separate grasp/trajectory stage.
$^{\S}$Implicit interaction: interaction encouraged only via latent/objective terms (no explicit contact signal).
$^{\P}$Explicit contact: an explicit contact representation is predicted/used for generation.
}
\label{tab:component_comparison}
\end{table}

Joint generation, however, does not guarantee that the synthesized motion will faithfully satisfy the predicted contacts at inference.
We therefore introduce Contact Inner Guidance (CIG), an inference-time sampling strategy that steers the denoising process toward motions that better respect hand--object proximity. 
Concretely, CIG enforces consistency between the predicted contact maps and the contact inferred from the synthesized geometry, and uses classifier-based guidance to steer hand motion, reducing penetration and floating artifacts.
Crucially, CIG operates strictly as a training-free guidance mechanism during inference without introducing external neural networks, separate grasp planning stages, or post-processing refinement steps, thereby fully preserving the unified single-stage nature of our pipeline.
Extensive experiments show that \textsc{JointHOI} synthesizes more natural motions with better text adherence, and that contact guidance further improves physical plausibility by reducing penetration and floating.
We also conduct targeted ablations to isolate each component and validate the overall design.

Our primary contributions are summarized as follows:
\begin{itemize}[topsep=0pt,itemsep=0pt,parsep=0pt,partopsep=0pt]
    \item We introduce \textbf{\textsc{JointHOI}}, a single-stage joint diffusion framework that generates 3D hand--object motion together with \emph{dynamic, distance-based} contact maps directly from text.
    \item We present \textbf{Contact Inner Guidance (CIG)}, an inference-time guided sampling strategy with proximity-aware weighting that leverages co-generated contact maps to steer denoising and reduce penetration and floating artifacts without requiring secondary training stages.
    \item We provide extensive evaluations and targeted ablations demonstrating improved text adherence, motion naturalness, and physical plausibility, validating the effectiveness of each component and the overall pipeline.
\end{itemize}

\section{Related Works}
\label{sec:related_works}


\noindent\textbf{Text to Hand--Object Interaction Generation.} 
Recent methods generate HOI sequences conditioned on text and object geometry to enable semantic control without explicit object trajectories, but realistic HOI requires fine-grained, temporally evolving contact that is hard to infer from motion alone.
Text2HOI~\cite{cha2024text2hoi} and DiffH2O~\cite{christen2024diffh2o} factorize generation into multi-stage designs (\eg, grasp or contact map prediction followed by motion refinement), while LatentHOI~\cite{Muchen_LatentHOI} and HOI-GPT~\cite{hoigpt} employ latent priors or autoregressive modeling. 
While similar contact-aware or geometry-guided representations have recently emerged in the broader human--object interaction domain~\cite{xue2025guiding,li2026conta}, existing hand--object generation frameworks still heavily rely on complex multi-stage pipelines or static binary cues, which hinder modeling long-horizon, time-varying contact. 
In contrast, \textsc{JointHOI} performs single-stage generation and explicitly models interaction with \emph{distance-based, dynamic} contact maps, providing a spatiotemporal contact signal that can be enforced during sampling.


\smallskip
\noindent\textbf{Contact in Human Motion Modeling.} 
Contact is widely used to represent interactions between 3D human surfaces and environments~\cite{affordmotion,cghoi}, but these full-body methods lack the fine-grained hand articulation required for realistic HOI, which is emphasized in grasp generation~\cite{contactgen,graspTTA,nl2contact}. 
In hand--object reconstruction, frameworks like ContactOpt~\cite{grady2021contactopt} and Cao \etal~\cite{cao2021reconstructing} utilize contact optimization to resolve physical violations from vision inputs. 
Similarly, for motion generation, HOIDINI~\cite{hoidini} and CODA~\cite{coda} incorporate contact constraints via inference-time optimization (\eg, DNO~\cite{karunratanakul2023dno}), often requiring hundreds of steps. 
In contrast, we learn bimanual interactions within a single diffusion model and leverage co-generated contact for highly efficient inference-time guidance. 
Closest to our setting, BimArt~\cite{bimart} predicts HOI contact maps but assumes fixed object trajectories; we target \emph{text-to-HOI} where object motion is synthesized jointly with contact and hand motion.

\smallskip
\noindent{\textbf{Joint Generative Modeling and Inner Modalities.}}
Joint generation of multiple modalities has been increasingly explored both for multimodal synthesis and for improving the fidelity of a primary output.
Prior work co-generates appearance with geometry cues such as depth or 3D structure~\cite{zhang2025world,hassan2025gem,Byung-Ki_2025_ICCV,krishnan2025orchid} and jointly synthesizes audio with motion/visual signals~\cite{kwon2025jamflowjointaudiomotionsynthesis}.
General any-to-any frameworks~\cite{UIO,UIO2,yu2023scalingautoregressivemultimodalmodels} and masked multimodal modeling (\eg, 4M)~\cite{4m21,mizrahi20234m} further suggest the benefit of shared representations across modalities.
More recently, auxiliary \emph{inner} modalities -- signals directly derivable from the primary output -- have been used as additional generation targets to strengthen generation; for example, VideoJAM~\cite{chefer2025videojam} co-generates optical flow to enhance temporal coherence, and Redi~\cite{kouzelis2025boosting} co-generates DINOv2 features to improve image synthesis.
Motivated by the central role of contact in HOI, we treat dynamic contact maps as an inner modality of hand--object motion and leverage the co-generated contact signals to guide diffusion sampling toward physically plausible interactions.
%
\section{Method}
\label{sec:method}

We study text-driven generation of bimanual 3D hand--object interactions conditioned on object geometry.
The key challenge is \emph{contact-consistent} motion: small errors in contact timing or proximity readily produce artifacts such as penetration, jitter, and floating.
Prior text-to-HOI approaches often rely on multi-stage pipelines with static/binary contact cues or implicit latent priors, increasing complexity and limiting their ability to model \emph{time-varying} contact.

\begin{figure}[t!]
\centering
\includegraphics[width=\linewidth]{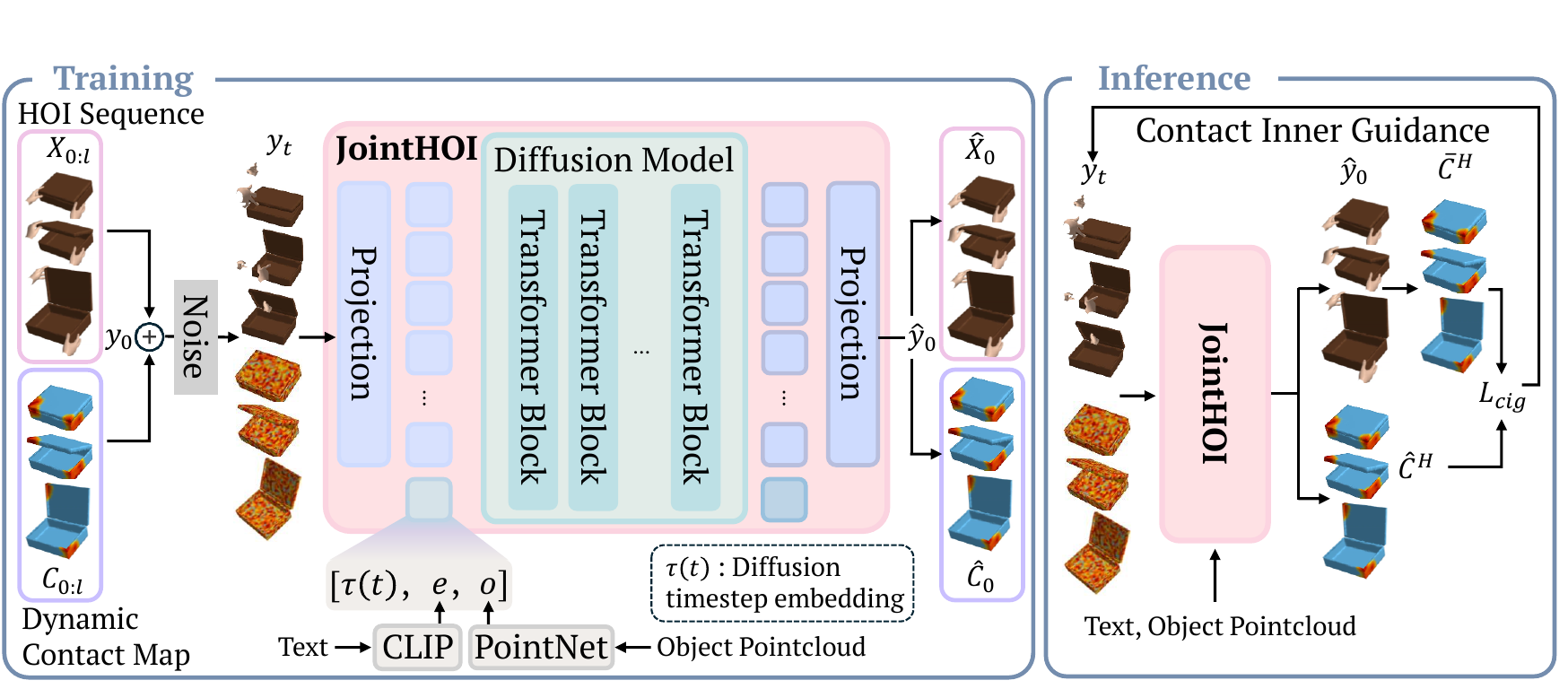}
\caption{\textbf{Overview of \textsc{JointHOI}.}
\textbf{Training:} We jointly diffuse an HOI sequence $y$ comprising bimanual motion, object motion, and per-hand dynamic contact maps, and train a Transformer diffusion model to predict $\hat{y}_0$ from noisy inputs $y_t$ conditioned on text (CLIP) and object geometry (PointNet).
\textbf{Inference:} We apply Contact Inner Guidance (CIG) using a contact-consistency energy $\mathcal{L}_{\mathrm{cig}}$ that encourages agreement between the generated contact maps and the contact implied by the synthesized geometry, improving physical plausibility without post-processing.
}
\label{fig:main_fig}
\end{figure}

We address these limitations with \textbf{\textsc{JointHOI}}, a \emph{single-stage joint diffusion} framework that jointly synthesizes HOI motion and a \emph{distance-based dynamic contact map} in one generative process (\cref{fig:main_fig}).
Our approach consists of (i) a continuous, temporally dense contact representation that explicitly tracks frame-wise hand--object proximity (Sec.~\ref{subsec:dynamic_contact_map}),
(ii) a joint diffusion model that learns contact--motion coupling in a shared latent space by generating motion and contact simultaneously (Sec.~\ref{subsec:joint_diffusion}),
and (iii) Contact Inner Guidance (CIG), an inference-time guided sampling strategy that enforces agreement between explicitly generated contacts and motion-implied contacts (Sec.~\ref{subsec:sampling}).

\subsection{Preliminaries}
\label{subsec:preliminaries}

\noindent\textbf{Notation.}
We use $t\in\{1,\dots,T\}$ to index diffusion steps of a \emph{noised} variable (\eg, $x_t$, $y_t$), and $l\in\{1,\dots,L_{\max}\}$ to index temporal frames (\eg, $x_l$, $y_l$).
We omit the diffusion index when referring to clean motion variables and deterministic geometry mappings, and write $(\cdot)_t$ only for noised diffusion variables.

\smallskip\noindent\textbf{Diffusion models.}
Diffusion models learn to invert a fixed Markov noising process.
Given clean data $x_0$, the forward process produces $\{x_t\}_{t=1}^T$ via
\begin{equation}
q(x_t \mid x_{t-1})=\mathcal{N}\!\big(x_t;\sqrt{1-\beta_t}\,x_{t-1},\,\beta_t I\big),
\label{eq:diffusion_forward}
\end{equation}
which yields the marginal
\begin{equation}
x_t=\sqrt{\bar{\alpha}_t}\,x_0+\sqrt{1-\bar{\alpha}_t}\,\epsilon,
\qquad
\epsilon\sim\mathcal{N}(0,I),\ \ 
\bar{\alpha}_t=\prod_{s=1}^t(1-\beta_s).
\label{eq:diffusion_marginal}
\end{equation}
The reverse process is parameterized by a denoiser $D_\theta$ conditioned on an external signal $c$ (\eg, text/object) and the timestep.
We adopt \emph{clean-sample prediction} and estimate
\begin{equation}
\hat{x}_0=D_\theta(x_t,c,t),
\label{eq:clean_prediction}
\end{equation}
trained with the reconstruction objective
\begin{equation}
\mathcal{L}_{\mathrm{rec}}
=
\mathbb{E}_{x_0,t,\epsilon}
\left[\left\lVert x_0 - D_\theta(x_t,c,t)\right\rVert_2^2\right].
\label{eq:rec_loss}
\end{equation}
At inference time, we sample $x_T\sim\mathcal{N}(0,I)$ and iteratively apply the learned reverse transitions to obtain a final sample.

\smallskip\noindent\textbf{3D hand--object motion representation.}
We represent a bimanual interaction as three time-aligned sequences
\begin{equation}
x=\big(x^L,x^R,x^O\big)
\quad \text{of length } L_{\max},
\label{eq:hoi_sequence}
\end{equation}
where
$x^L=\{x^L_{l}\}_{l=1}^{L_{\max}}$, $x^R=\{x^R_{l}\}_{l=1}^{L_{\max}}$, and $x^O=\{x^O_{l}\}_{l=1}^{L_{\max}}$ share the same frame index $l$.
Each hand state is $x^H_{l}=[t^H_l,\theta^H_l]\in\mathbb{R}^{99}$ for $H\in\{L,R\}$,
where $t^H_l\in\mathbb{R}^3$ is the global translation and
$\theta^H_l\in\mathbb{R}^{16\times 6}$ are MANO pose parameters in 6D rotation form (flattened).
The object state is $x^O_{l}=[t^O_l,\varphi^O_l,\delta_l]\in\mathbb{R}^{10}$, where
$t^O_l\in\mathbb{R}^3$ is translation, $\varphi^O_l\in\mathbb{R}^6$ is 6D rotation, and $\delta_l\in\mathbb{R}$ is an articulation scalar.
For rigid-object datasets (\eg, GRAB~\cite{GRAB}), we omit articulation and use $x^O_{l}=[t^O_l,\varphi^O_l]\in\mathbb{R}^9$.
Our model outputs predicted trajectories $\hat{x}=\big(\hat{x}^L,\hat{x}^R,\hat{x}^O\big)$; quantities derived from predictions inherit the hat notation.

\smallskip\noindent\textbf{Differentiable geometry.}
We use MANO~\cite{MANO} as a differentiable parametric hand model to obtain per-frame meshes and joints:
\begin{equation}
(V_H,J_H)=\mathrm{MANO}(x^H),
\label{eq:mano}
\end{equation}
where
$V_H\in\mathbb{R}^{L_{\max}\times V\times 3}$ and $J_H\in\mathbb{R}^{L_{\max}\times J\times 3}$ ($V{=}778$, $J{=}21$).
For the object, given a canonical point cloud $P\in\mathbb{R}^{N\times 3}$, we map it to the world frame at each time $l$ using the object state:
\begin{equation}
P^{\,l}_{\mathrm{def}}=\mathcal{T}(P;t^O_l,\varphi^O_l,\delta_l),
\label{eq:object_transform}
\end{equation}
where $\mathcal{T}$ applies per-frame translation/rotation (and articulation $\delta_l$ when available; omitted for rigid objects), yielding $P_{\mathrm{def}}\in\mathbb{R}^{L_{\max}\times N\times 3}$ (and analogously $\hat{P}_{\mathrm{def}}$ from $\hat{x}^O$).

\subsection{Dynamic Contact Map}
\label{subsec:dynamic_contact_map}

\begin{figure}[t]
  \centering
  \includegraphics[width=\linewidth]{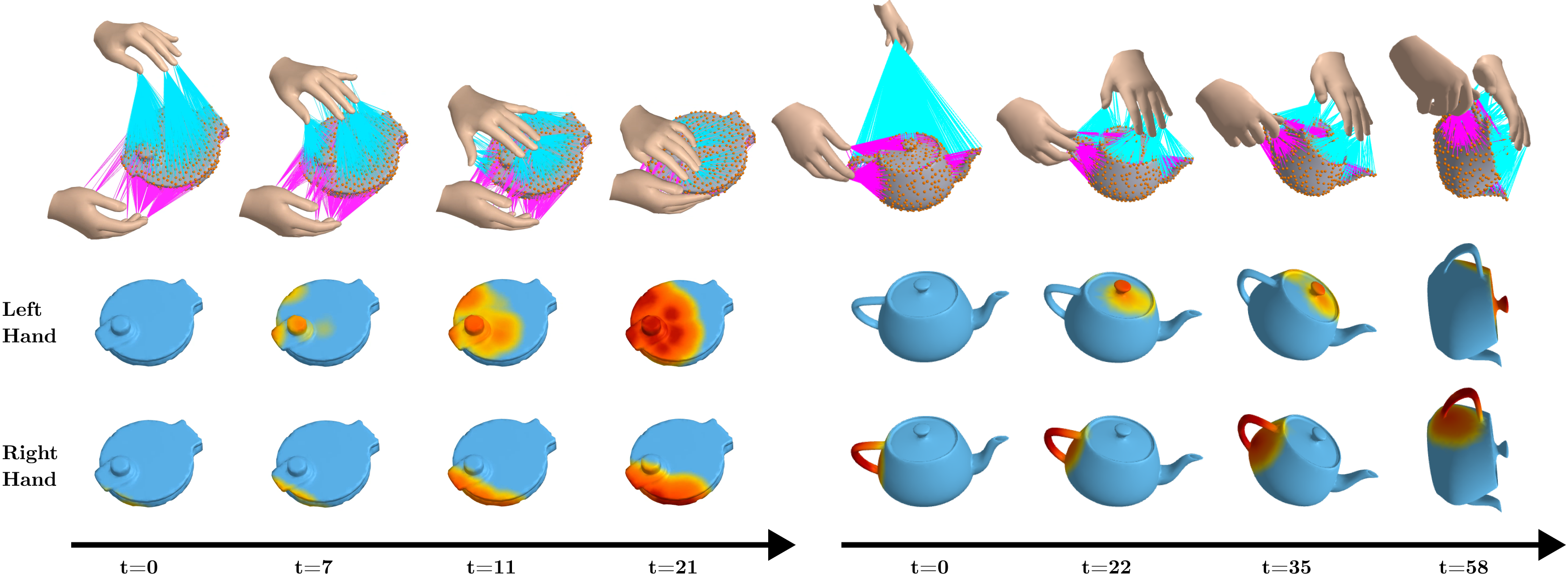}
\caption{\textbf{Dynamic contact map visualization.}
\textbf{Top:} For each object-surface anchor, we connect it to its nearest point on the left (cyan) and right (magenta) hand, illustrating the per-hand proximity field over time.
\textbf{Bottom:} We render the corresponding anchor-to-hand distances as heatmaps on the object surface (shown separately for left/right hands), highlighting how contact regions emerge, move, and disappear throughout the interaction.}
  \label{fig:contact_map_vis}
\end{figure}

Prior text-to-HOI methods either use \emph{static} contact cues (\eg, a binary map) as a conditioning signal or learn an \emph{implicit} interaction prior in a latent space (\eg, via a VAE~\cite{kingma2013auto}).
In contrast, we introduce a \emph{dynamic contact map} that is (i) \emph{temporally dense}, capturing frame-wise contact evolution, and (ii) \emph{explicit}, providing an interpretable geometric signal that can be directly leveraged at inference time.
We represent contact as \emph{continuous distances} rather than binary labels, which avoids threshold sensitivity and yields a smooth signal near contact boundaries -- particularly beneficial for gradient-based guidance during sampling.
As illustrated in \cref{fig:contact_map_vis}, this distance field yields a temporally evolving proximity signal over the object surface for each hand.

\smallskip\noindent\textbf{Anchor-based distance representation.}
For each object, we predefine a fixed set of $N_c{=}1024$ anchor points
\begin{equation}
\mathcal{A}=\{a_n\}_{n=1}^{N_c}
\label{eq:anchors}
\end{equation}
in the object canonical frame.
We employ farthest point sampling (FPS) to ensure that $\mathcal{A}$ uniformly covers the object surface.
These anchors define an object-centric discretization that is shared across time and invariant to global motion, allowing contacts to be tracked consistently as the object moves.
Given an object motion sequence $x^O=\{x^O_l\}_{l=1}^{L_{\max}}$, we transform anchors into the global frame at each time $l$, yielding
\begin{equation}
\mathcal{A}_l=\{a_{l,n}\}_{n=1}^{N_c}.
\label{eq:anchors_global}
\end{equation}
We define bimanual contact at frame $l$ as $C_l=(C^L_l,C^R_l)$ with
$C^H_l\in\mathbb{R}^{N_c}$ for $H\in\{L,R\}$, where each entry stores the minimum Euclidean distance from an anchor point to the corresponding hand surface.
In practice, we compute this distance by taking the minimum over hand mesh vertices, which provides an efficient and differentiable approximation of surface proximity.
Let $\mathcal{U}^H_l=\{V_H[l,m,:]\}_{m=1}^{V}$ denote the hand mesh vertices at frame $l$; we compute
\begin{equation}
C^H_l[n] \;=\; \min_{u\in \mathcal{U}^H_l} \|u - a_{l,n}\|_2,
\qquad n=1,\dots,N_c,\ \ H\in\{L,R\}.
\label{eq:contact_distance}
\end{equation}

\smallskip\noindent\textbf{Design choices.}
As shown in \cref{fig:contact_map_vis}, contact patterns are asymmetric across hands and evolve over time with complex spatial patterns.
Therefore, we use per-hand bimanual contact maps~\cite{bimart} to reduce ambiguity in bimanual interactions, and adopt dynamic, distance-based maps to represent complex spatiotemporal contact structures.


\subsection{Joint Diffusion Model}
\label{subsec:joint_diffusion}

Rather than treating contact as a separate, precomputed condition, we \emph{jointly} generate HOI motion and contact dynamics with a single diffusion model.
Given an object geometry descriptor and a text prompt $c$, we directly model the joint distribution of bimanual hand motion, object trajectory, and dynamic contact.
This joint formulation reduces error propagation inherent to sequential pipelines and enables the denoiser to learn \emph{contact--motion coupling} in a shared latent space.
Intuitively, motion and contact must co-vary under physical constraints: as the hand approaches and reorients around a surface patch, contact distances should decrease smoothly and evolve consistently over time.

\smallskip\noindent\textbf{Unified sequence representation.}
Let $x^L_l\in\mathbb{R}^{99}$, $x^R_l\in\mathbb{R}^{99}$, and $x^O_l\in\mathbb{R}^{10}$ denote per-frame motion states (Sec.~\ref{subsec:preliminaries}), and let $C^L_l,C^R_l\in\mathbb{R}^{N_c}$ denote per-frame dynamic contact maps (Sec.~\ref{subsec:dynamic_contact_map}).
We concatenate them into a single per-frame token,
\begin{equation}
y_l \;=\; \big[x^L_l,\ x^R_l,\ x^O_l,\ C^L_l,\ C^R_l\big]\in\mathbb{R}^{d_y},
\qquad
d_y = 99+99+10+2N_c,
\label{eq:joint_token}
\end{equation}
and stack over time to obtain the joint sequence
\begin{equation}
y=\{y_l\}_{l=1}^{L_{\max}}\in\mathbb{R}^{L_{\max}\times d_y}.
\label{eq:joint_sequence}
\end{equation}
We define diffusion directly on $y$, so each transformer layer can model both (i) \emph{within-frame} dependencies between kinematics and contact and (ii) \emph{across-frame} temporal evolution via self-attention.

\smallskip\noindent\textbf{Transformer denoiser and conditioning.}
We use a transformer denoiser that takes the noisy sequence $y_t$ and predicts the corresponding clean sequence $\hat{y}_0$.
The model is conditioned on (i) an object feature vector $\mathbf{o}\in\mathbb{R}^{1088}$ (global/local PointNet~\cite{Qi_2017_CVPR} features, scale, centroid), (ii) a CLIP text embedding $\mathbf{e}\in\mathbb{R}^{d_e}$, and (iii) a timestep embedding $\tau(t)$.
We inject these conditions using \emph{prefix conditioning}: we append a small set of learnable condition tokens to the beginning of the token sequence,
$z=[z^{\text{text}},z^{\text{obj}},z^{t}]$, initialized from $\mathbf{e}$, $\mathbf{o}$, and $\tau(t)$, respectively.
The denoiser then applies standard \emph{full self-attention} over the concatenated sequence $[z;\ \phi(y_t)]$, where $\phi(\cdot)$ linearly projects $y_t$ to the model dimension.
As a result, each frame token can attend to the text token to capture action semantics and to the object token to adapt contact evolution to object-specific geometry (\eg, different grasp regions for a mug handle vs.\ a cup body), without introducing a separate cross-attention module.
We write the denoiser compactly as
\begin{equation}
\hat{y}_0 \;=\; D_\theta\!\big(y_t,\ \mathbf{o},\ \mathbf{e},\ \tau(t)\big).
\label{eq:joint_denoiser}
\end{equation}

\noindent\textbf{Training objective.}
We adopt clean-sample prediction and train the denoiser to reconstruct the clean joint sequence:
\begin{equation}
\mathcal{L}_{\mathrm{joint}}
=
\mathbb{E}_{y,t,\epsilon}\Big[\big\|y - D_\theta(y_t,\ \mathbf{o},\ \mathbf{e},\ \tau(t))\big\|_2^2\Big].
\label{eq:joint_loss}
\end{equation}
Because $y$ contains both kinematic variables and contact distances, optimizing $\mathcal{L}_{\mathrm{joint}}$ encourages the transformer to learn \emph{cross-modal dependencies} through self-attention:
motion dimensions can inform contact dimensions and vice versa, allowing the model to internalize physically grounded correlations such as decreasing distances during approach, increasing distances during release, and coordinated contact changes during sliding or rolling.

\smallskip\noindent\textbf{Inference and parsing.}
At inference time, we sample $y_T\sim\mathcal{N}(0,I)$ and iteratively apply the learned reverse transitions conditioned on $(\mathbf{o},\mathbf{e})$ to obtain $\hat{y}_0$.
We parse $\hat{y}_0$ into $\hat{x}^L,\hat{x}^R,\hat{x}^O,\hat{C}^L,\hat{C}^R$ by slicing the corresponding dimensions, and compute hand meshes and motion-implied contact geometry using the operators in Sec.~\ref{subsec:preliminaries} and Sec.~\ref{subsec:dynamic_contact_map}.
\subsection{Contact Inner Guidance}
\label{subsec:sampling}

Although our model is trained to jointly predict motion and contact, long and highly dynamic interactions can still accumulate geometric inconsistencies during iterative denoising.
We therefore improve physical plausibility at inference time by enforcing consistency between
(i) the contact maps explicitly generated by the diffusion model and
(ii) the contact maps implied by the generated hand--object geometry.
We achieve this via a classifier-based \emph{guided sampling} scheme, where a differentiable contact-consistency energy provides gradients that steer the denoising trajectory.
We refer to this procedure as \emph{Contact Inner Guidance (CIG)}, since it leverages contact -- a co-generated inner modality -- to guide hand--object motion generation at inference time.

\smallskip\noindent\textbf{Contact consistency energy.}
At a reverse step, let $\hat{y}_0$ be the current clean estimate predicted by the denoiser.
We parse $\hat{y}_0$ into $\hat{x}^L,\hat{x}^R,\hat{x}^O,\hat{C}^L,\hat{C}^R$.
From the predicted motion, we compute a \emph{motion-implied} dynamic contact map
$\bar{C}^H=\{\bar{C}^H_l\}_{l=1}^{L_{\max}}$ using the same construction as Sec.~\ref{subsec:dynamic_contact_map}:
we transform anchors with $\hat{x}^O$ to obtain $\hat{\mathcal{A}}_l$,
extract hand vertices $\hat{\mathcal{U}}^H_l$ from $\mathrm{MANO}(\hat{x}^H)$,
and set $\bar{C}^H_l[n]=\min_{u\in \hat{\mathcal{U}}^H_l}\|u-\hat{a}_{l,n}\|_2$.
We measure agreement between the explicitly generated contacts and the motion-implied contacts with the log-ratio energy
\begin{equation}
\mathcal{L}_{\mathrm{cig}}
=
\sum_{H\in\{L,R\}}
\left\|
\log(\hat{C}^H + \varepsilon)
-
\log(\bar{C}^H + \varepsilon)
\right\|_1,
\label{eq:cig_energy}
\end{equation}
where $\hat{C}^H,\bar{C}^H\in\mathbb{R}^{L_{\max}\times N_c}$ stack per-frame contact vectors. We use the log scale to emphasize fine-grained, near-contact regions (small hand--object distances), motivated by the observation that these regions carry the most critical cues for realistic interaction, whereas far-field distances are less informative. The constant $\varepsilon$ stabilizes the computation near zero.

\smallskip\noindent\textbf{Classifier-based guidance.}
We guide sampling by nudging the noisy variable toward samples that reduce $\mathcal{L}_{\mathrm{cig}}$.
Concretely, at each reverse step we first obtain the model prediction
$\hat{y}_0=D_\theta(y_t,\mathbf{o},\mathbf{e},\tau(t))$,
compute $\mathcal{L}_{\mathrm{cig}}(\hat{y}_0)$, and backpropagate through the differentiable geometry operators (MANO and the distance computation) to obtain a guidance gradient with respect to the current noisy variable, $\nabla_{y_t}\mathcal{L}_{\mathrm{cig}}$.
We then apply a guidance update with scale $w$,
\begin{equation}
y_t \leftarrow y_t - w\, \nabla_{y_t}\mathcal{L}_{\mathrm{cig}},
\label{eq:cig_update}
\end{equation}
followed by the standard diffusion reverse transition.
Intuitively, this steers denoising toward joint samples whose explicitly generated contact distances match those physically induced by the generated motion, reducing artifacts such as contact jitter, floating, and interpenetration.

\section{Experiments}
\label{sec:experiments}

\subsection{Datasets}
\label{subsec:dataset}
We evaluate \textsc{JointHOI} on two widely used bimanual hand--object interaction datasets, GRAB~\cite{GRAB} and ARCTIC~\cite{arctic}.
For both datasets, we use the text prompts released by Text2HOI~\cite{cha2024text2hoi} as language conditioning.
Since no official train/test split is publicly provided for these datasets in the text-to-HOI setting, we construct the splits using a shared protocol:
we form a balanced test set by uniformly sampling object--action pairs, and use the remaining sequences for training.
All methods are trained and evaluated on the same splits.

\smallskip
\noindent\textbf{GRAB~\cite{GRAB}.}
It provides bimanual interactions with 51 rigid objects across 29 action categories.
All sequences are downsampled to 30\,fps, and interaction clips are padded or truncated to a maximum length of 196 frames.

\noindent\textbf{ARCTIC~\cite{arctic}.}
It contains bimanual interactions with 11 articulated objects spanning 10 action categories.
In addition to motion, it provides object articulation angles, enabling evaluation on interactions involving moving parts.
All sequences are downsampled to 30\,fps and padded or truncated to a maximum length of 64 frames.

\subsection{Implementation Details}
\label{subsec:implementation}
All experiments are conducted on a single NVIDIA RTX A6000 GPU (48GB).
We optimize the network using Adam ($lr{=}10^{-4}$) and a linear noise schedule with $T{=}1{,}000$ diffusion steps for training, and use 100 sampling steps at inference.
We train with batch size 128 on GRAB and 64 on ARCTIC.
Training is run for up to 380K iterations on GRAB and 60K iterations on ARCTIC, with early stopping when the training loss plateaus.
At inference time, we generate sequences with the same number of frames as the corresponding ground-truth clip to ensure consistent comparison across metrics.

\smallskip
\noindent\textbf{Baselines.}
We compare \textsc{JointHOI} to state-of-the-art text-to-HOI methods, including MDM~\cite{tevet2023human}, DiffH2O~\cite{christen2024diffh2o}, LatentHOI~\cite{Muchen_LatentHOI}, and Text2HOI~\cite{cha2024text2hoi}.
All baselines are trained using their official implementations under the same data splits and evaluation protocol.
MDM serves as a single-stage diffusion baseline without interaction-specific guidance.
For the two-stage DiffH2O, we report its standard setting and DiffH2O$^{*}$, where the grasp stage is conditioned on ground-truth grasp signals to isolate the quality of interaction synthesis.
Similarly, we report Text2HOI and Text2HOI$^{*}$, where Text2HOI$^{*}$ removes the post-hoc refinement module to evaluate the raw motion generated from contact-map conditioning.

\subsection{Evaluation Metrics}
\label{subsec:eval_metric}
We evaluate \textsc{JointHOI} from two complementary perspectives: (i) semantic motion quality and (ii) physical plausibility of the hand--object interaction.

\smallskip
\noindent\textbf{Motion Metrics.}
Following the protocol of IMOS~\cite{ghosh2023imos}, we train an RNN-based action classifier for each dataset and use its hidden features as motion embeddings.
We report:
\begin{itemize}[topsep=0pt,itemsep=0pt,parsep=0pt,partopsep=0pt]
    \item \textbf{Accuracy (Acc):} Top-1 and Top-3 classification accuracy of generated motions under the action classifier, measuring semantic alignment with the target action.
    \item \textbf{FID:} Fr\'echet distance between feature distributions of real and generated sequences, quantifying overall motion realism.
\end{itemize}

\noindent\textbf{Physical Plausibility Metrics.}
Since motion metrics may miss contact artifacts, we further evaluate physical consistency by averaging the following metrics over both hands:
\begin{itemize}[topsep=0pt,itemsep=0pt,parsep=0pt,partopsep=0pt]
    \item \textbf{Interpenetration Volume (IV) and Depth (ID):}
    Following LatentHOI~\cite{Muchen_LatentHOI}, we measure the intersection volume ($\text{cm}^3$) and maximum penetration depth ($\text{cm}$) between hand and object meshes.
    Lower values indicate fewer physical violations.
    \item \textbf{Contact Ratio (CR):}
    Following Bimart~\cite{bimart}, CR is the fraction of frames (where the object translates or articulates) that maintain valid contact (distance $<5$\,mm).
    Since CR can be trivially increased by penetration, we consider the best interaction to have \emph{CR close to GT} while keeping IV/ID low.
\end{itemize}

\subsection{Experimental Results}
\label{subsec:results}

\noindent\textbf{Main results on ARCTIC and GRAB.}
\cref{tab:comparison_results} summarizes quantitative results on ARCTIC and GRAB.
On \textbf{ARCTIC}, \textsc{JointHOI} achieves the highest semantic fidelity and the lowest FID among synthesized methods, while substantially reducing penetration and keeping CR close to GT. It consistently outperforms multi-stage baselines and remains highly competitive even against the oracle-assisted DiffH2O$^{*}$.
On \textbf{GRAB}, \textsc{JointHOI} delivers the best Top-3 accuracy and lowest FID, demonstrating superior motion realism. While Text2HOI shows a slightly higher Top-1 accuracy, our significant Top-3 gain indicates that \textsc{JointHOI} more reliably captures ambiguous action semantics (\eg, \emph{use} vs. \emph{inspect}). Notably, \textsc{JointHOI} achieves this while halving the penetration volume (IV) compared to Text2HOI. 
These results highlight that jointly modeling motion and interaction constraints leads to higher-fidelity, physically grounded interactions.

\begin{table}[t!]
\centering
\footnotesize
\caption{
Quantitative comparison on ARCTIC and GRAB.
Bold indicates the best performance among synthesized methods (excluding GT).
$^{*}$ denotes the oracle-assisted variant where the grasp stage is conditioned on ground-truth grasp signals.
}
\label{tab:comparison_results}
\setlength{\tabcolsep}{5pt}
\renewcommand{\arraystretch}{1.1}
\resizebox{\textwidth}{!}{%
\begin{tabular}{l|l|cc|c|cc|c}
\toprule
\multirow{2}{*}{\textbf{Dataset}} &
\multirow{2}{*}{\textbf{Method}} &
\multicolumn{2}{c|}{\textbf{Acc} $\uparrow$} &
\multirow{2}{*}{\textbf{FID} $\downarrow$} &
\multicolumn{2}{c|}{\textbf{Interpenetration} $\downarrow$} &
\multirow{2}{*}{\makecell{\textbf{CR} (\%) $\rightarrow$}} \\
& &
\makecell{Top-1} & \makecell{Top-3} &
& \makecell{IV (cm$^3$)} & \makecell{ID (cm)} & \\
\midrule
\multirow{8}{*}{\textbf{ARCTIC}}
& GT                 & 0.966 & 0.992 & --    & 4.419 & 0.288 & 95.56 \\
\cmidrule(lr){2-8}
& MDM~\cite{tevet2023human}                & 0.739 & 0.839 & 0.365 & 9.542 & 0.581 & 61.37 \\
& DiffH2O$^{*}$ \cite{christen2024diffh2o} & 0.636 & 0.862 & 0.424 & 6.491 & 0.475 & 94.28 \\
& DiffH2O \cite{christen2024diffh2o}       & 0.560 & 0.823 & 0.547 & 6.412 & 0.527 & \textbf{94.91} \\
& LatentHOI \cite{Muchen_LatentHOI}        & 0.549 & 0.846 & 0.301 & 8.597 & 0.496 & 78.81 \\
& Text2HOI$^{*}$ \cite{cha2024text2hoi}    & 0.815 & 0.923 & 0.152 & 8.930 & 0.587 & 93.28 \\
& Text2HOI \cite{cha2024text2hoi}          & 0.816 & 0.927 & 0.148 & 8.281 & 0.524 & 97.33 \\
& \textbf{\textsc{JointHOI} (Ours)}                            & \textbf{0.948} & \textbf{0.983} & \textbf{0.033} & \textbf{4.406} & \textbf{0.426} & 93.71 \\
\midrule
\multirow{8}{*}{\textbf{GRAB}}
& GT                 & 0.779 & 0.895 & --    & 2.790 & 0.526 & 95.02 \\
\cmidrule(lr){2-8}
& MDM~\cite{tevet2023human}                      & 0.395 & 0.500 & 1.203 & 4.915 & 0.650 & 81.22 \\
& DiffH2O$^{*}$ \cite{christen2024diffh2o} & 0.595 & 0.737 & 0.410 & \textbf{3.282} & 0.610 & 87.24 \\
& DiffH2O \cite{christen2024diffh2o}       & 0.458 & 0.526 & 0.533 & 3.607 & 0.846 & 87.82 \\
& LatentHOI \cite{Muchen_LatentHOI}        & 0.584 & 0.721 & 0.214 & 4.356 & 0.570 & 88.39 \\
& Text2HOI$^{*}$ \cite{cha2024text2hoi}    & \textbf{0.716} & 0.805 & 0.118 & 8.114 & 0.810 & 97.85 \\
& Text2HOI \cite{cha2024text2hoi}          & 0.711 & 0.831 & 0.116 & 7.790 & 0.778 & 98.24 \\
& \textbf{\textsc{JointHOI} (Ours)}                            & 0.663 & \textbf{0.847} & \textbf{0.031} & 3.419 & \textbf{0.525} & \textbf{93.00} \\
\bottomrule
\end{tabular}%
}
\end{table}

\begin{table}[t!]
\centering
\footnotesize
\caption{Incremental ablation from the baseline by progressively adding our modules.
Each row adds one component on top of the previous setting.}
\label{tab:incremental_ablation}
\setlength{\tabcolsep}{4pt}
\renewcommand{\arraystretch}{1.1}
\resizebox{\textwidth}{!}{%
\begin{tabular}{l|cc|c|cc|c}
\toprule
\multirow{2}{*}{\textbf{Setting}} &
\multicolumn{2}{c|}{\textbf{Acc} $\uparrow$} &
\multirow{2}{*}{\textbf{FID} $\downarrow$} &
\multicolumn{2}{c|}{\textbf{Interpenetration} $\downarrow$} &
\multirow{2}{*}{\textbf{CR} (\%) $\rightarrow$} \\
& \makecell{Top-1} & \makecell{Top-3} &
& \makecell{IV (cm$^3$)} & \makecell{ID (cm)} & \\
\midrule
Baseline (two-stage)                       & 0.520 & 0.738 & 0.291 & 8.096 & 0.528 & 73.40 \\
\quad + One-stage (joint)                  & 0.795 & 0.919 & 0.274 & 7.057 & 0.548 & 81.72 \\
\quad + Dynamic contact map (unified)      & 0.894 & 0.948 & 0.145 & 7.146 & 0.432 & 85.28 \\
\quad + Dynamic contact map (L/R)          & 0.935 & 0.981 & 0.038 & 6.988 & 0.439 & 92.21 \\
\quad + Contact Inner Guidance            & \textbf{0.948} & \textbf{0.983} & \textbf{0.033} & \textbf{4.406} & \textbf{0.426} & \textbf{93.71} \\
\bottomrule
\end{tabular}%
}
\end{table}

\smallskip
\noindent\textbf{Effect of our design choices.}
\cref{tab:incremental_ablation} reports an incremental ablation on ARCTIC.
Switching from a two-stage baseline to one-stage joint training already yields a large improvement in semantic fidelity (Top-1 Acc $0.520\!\rightarrow\!0.795$), supporting our motivation that joint optimization mitigates stage-wise error accumulation.
Introducing a distance-based dynamic contact map further improves realism (FID $0.274\!\rightarrow\!0.145$), consistent with the intuition that contact should be modeled as a time-varying signal rather than a static cue.
Notably, separating the dynamic contact map by hand (L/R) provides a substantial additional gain in motion realism (FID $0.145\!\rightarrow\!0.038$), suggesting that bimanual interactions exhibit asymmetric, loosely coupled contact dynamics.
Finally, applying CIG achieves the best overall performance, markedly reducing penetration (IV $6.988\!\rightarrow\!4.406$) while maintaining strong motion quality.

\begin{figure}[t]
\centering
\includegraphics[width=\linewidth]{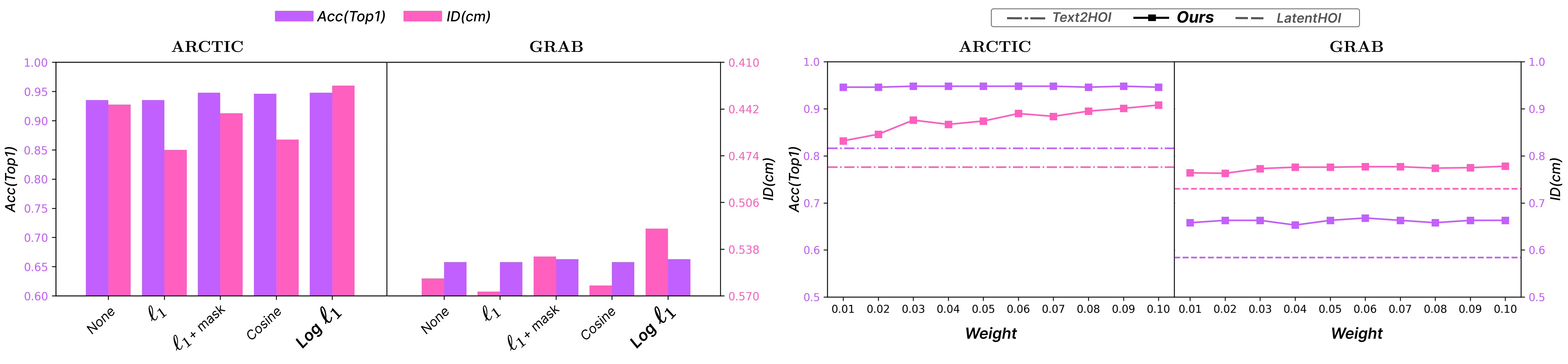}
\caption{\textbf{Ablation of Contact Inner Guidance (CIG).}
We report Top-1 action accuracy (Acc, left y-axis) and penetration depth (ID, right y-axis) on ARCTIC and GRAB; the ID axis is inverted, so higher is better for both metrics.
\textbf{Left:} Effect of contact-consistency loss design (None, $\ell_1$, $\ell_1{+}$mask, cosine, and Log-$\ell_1$).
Log-$\ell_1$ provides the best overall trade-off, consistently improving physical plausibility without sacrificing semantic accuracy.
\textbf{Right:} Effect of the guidance weight $w$.
Dashed horizontal lines denote the best prior (SOTA) results; CIG remains robust across a wide range of $w$ and consistently outperforms these baselines.}
\label{fig:cig_loss_compare}
\end{figure}

\smallskip
\noindent\textbf{Effect of CIG loss design and guidance weight.}
\cref{fig:cig_loss_compare} evaluates (left) different CIG consistency losses and (right) the guidance weight $w$, using Top-1 Acc and penetration depth (ID).
Across both datasets, the results support our proximity-based intuition: losses that emphasize near-contact discrepancies improve physical plausibility without hurting semantics.
In particular, Log-$\ell_1$ yields the best overall Acc--ID trade-off by focusing on relative errors near contact, but other proximity-aware variants (\eg, $\ell_1{+}$mask) also perform strongly, achieving competitive accuracy with reduced penetration.
The weight sweep further shows that CIG is not sensitive to $w$: ARCTIC and GRAB exhibit a broad stable range where Acc remains nearly unchanged while ID is consistently reduced.
We therefore use $w{=}0.05$ in all subsequent experiments.

\begin{table}[t]
\centering
\scriptsize
\caption{Inference time for a 196-frame sequence on an RTX A6000.}
\label{tab:inference_time}
\setlength{\tabcolsep}{4pt}
\renewcommand{\arraystretch}{1.05}
\resizebox{\columnwidth}{!}{%
\begin{tabular}{lccccc}
\toprule
\textbf{Metric} & MDM~\cite{tevet2023human} & DiffH2O~\cite{christen2024diffh2o} & LatentHOI~\cite{Muchen_LatentHOI} & Text2HOI~\cite{cha2024text2hoi} & \textsc{JointHOI} \\
\midrule
Time (s) $\downarrow$ & 10.61 & 124.44 & 15.95 & 16.79 & 10.98 \\
FPS $\uparrow$        & 18.47 & 1.57   & 12.29 & 11.67 & 17.85 \\
\bottomrule
\end{tabular}%
}
\end{table}

\smallskip
\noindent\textbf{Inference efficiency.}
Table~\ref{tab:inference_time} compares the inference time for a 196-frame sequence on an NVIDIA RTX A6000 GPU.
\textsc{JointHOI} is substantially more efficient than multi-stage or latent-based methods (DiffH2O~\cite{christen2024diffh2o}, LatentHOI~\cite{Muchen_LatentHOI}, and Text2HOI~\cite{cha2024text2hoi}), while remaining close to the single-stage baseline MDM~\cite{tevet2023human}.
The overhead of Contact Inner Guidance (CIG) is modest because it operates directly on the joint motion--contact representation using fixed anchors, preserving inference speed while improving physical plausibility.

\begin{figure}[t]
\centering
\includegraphics[width=\linewidth]{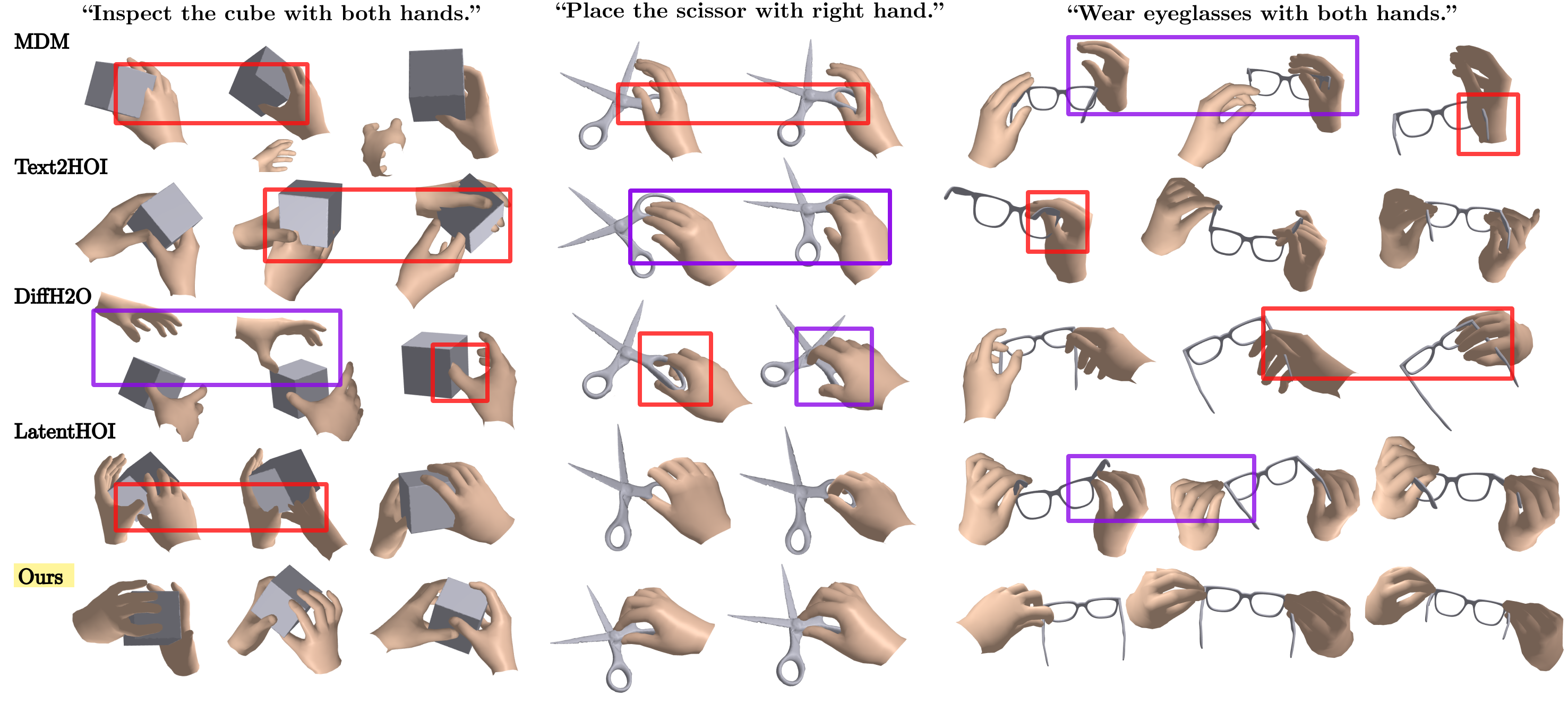}
\caption{\textbf{Qualitative comparison on text-to-HOI generation.}
We show representative frames for three prompts (columns) and compare MDM~\cite{tevet2023human}, Text2HOI~\cite{cha2024text2hoi}, DiffH2O~\cite{christen2024diffh2o}, and \textsc{JointHOI} (rows).
\textsc{JointHOI} produces more coherent bimanual coordination and more physically plausible hand--object interactions, with fewer artifacts such as floating (in purple boxes), unstable grasps, and penetration (in red boxes).}
\label{fig:qual:contact_map_vis}
\end{figure}

\smallskip
\noindent\textbf{Qualitative results.}
\cref{fig:qual:contact_map_vis} compares text-to-HOI generations from MDM, Text2HOI, DiffH2O, LatentHOI, and \textsc{JointHOI}.
Across prompts, \textsc{JointHOI} produces more coherent bimanual coordination and more stable hand--object proximity.
For \emph{``Inspect the cube with both hands''}, baselines often show weak two-hand engagement, while \textsc{JointHOI} maintains coordinated bimanual contact.
For \emph{``Place the scissor with right hand''}, competing methods frequently exhibit floating or unstable grasps during placement, whereas \textsc{JointHOI} preserves plausible right-hand control with fewer contact artifacts.
For \emph{``Wear an eyeglasses with both hands''}, a challenging case due to thin structures, \textsc{JointHOI} better aligns both hands to the frame and maintains consistent contact.
Overall, these results indicate that jointly modeling motion with dynamic contact and enforcing contact consistency during sampling yields more physically plausible interactions with fewer floating and interpenetration artifacts.

\begin{figure}[t]
  \centering
  \includegraphics[width=\linewidth]{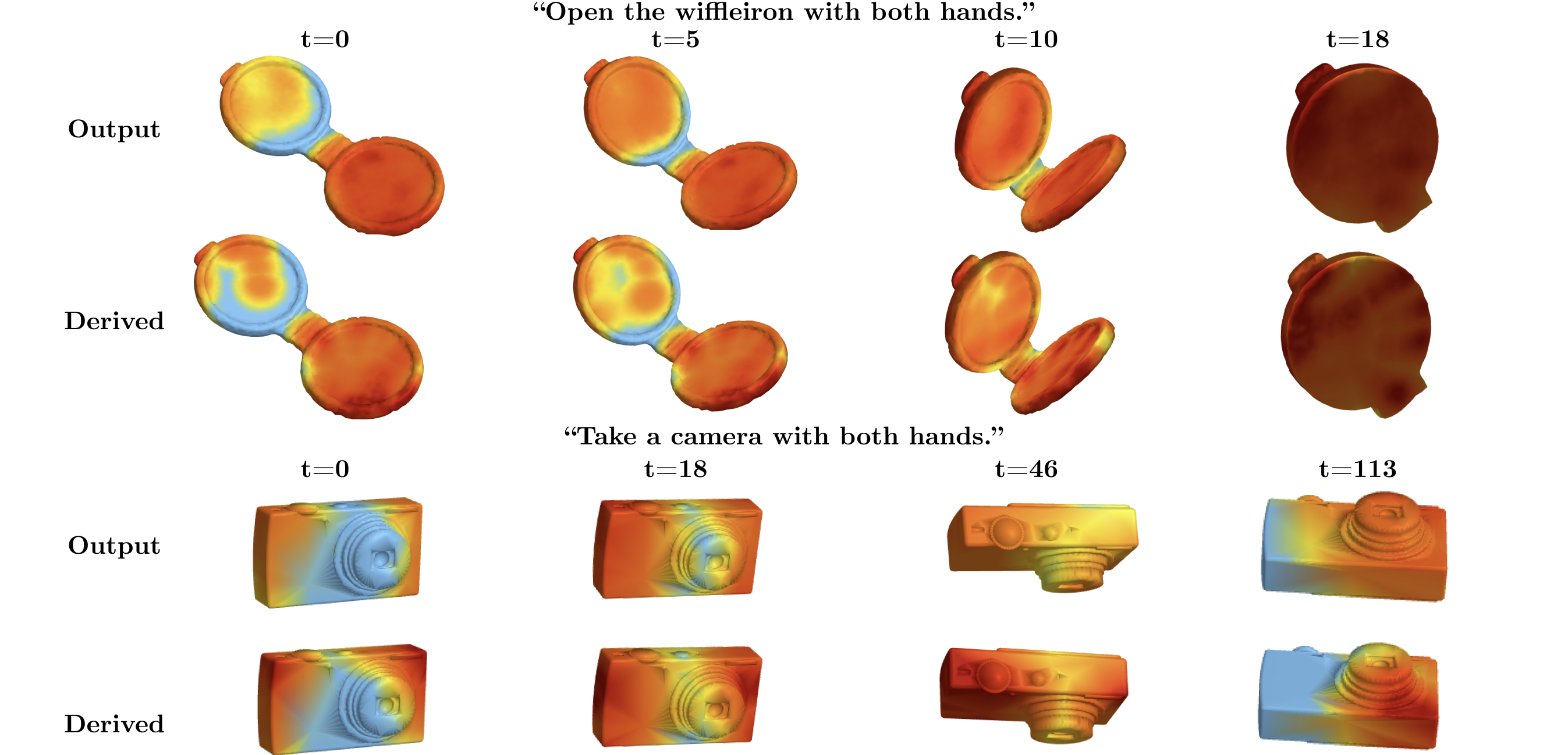}
  \caption{\textbf{Visualization of dynamic contact maps.}
  For each timestep, the first row (\textbf{Output}) shows the explicitly generated contact maps by \textsc{JointHOI}, and the second row (\textbf{Derived}) shows those analytically derived from the motions. The close agreement between them demonstrates that the predicted contact faithfully reflects the synthesized geometry.
  }
  \label{fig:raw_derived_vis}
\end{figure}

\smallskip
\noindent\textbf{Visualization of dynamic contact maps.}
\cref{fig:raw_derived_vis} compares the explicitly generated contact maps with those analytically derived from the synthesized hand--object motions.
The high consistency between the two demonstrates that our model effectively learns motion--contact coupling.
This supports our design choice of treating contact as an inner modality, showing that the predicted contact maps offer reliable signals for inference-time guidance.

\section{Conclusion}
\label{sec:conclusion}

We presented \textsc{JointHOI}, a single-stage joint diffusion framework for text-driven hand--object interaction generation that co-generates 3D motion with dynamic, distance-based contact maps.
By treating contact as an inner modality, \textsc{JointHOI} explicitly models the spatiotemporal evolution of hand--object proximity and learns contact--motion coupling within a unified generative process.
To further improve physical plausibility at inference time, we introduced contact-guided sampling via classifier-based guidance, which enforces consistency between the generated contact maps and the contact implied by the synthesized geometry, reducing common artifacts such as penetration, floating, and contact jitter.
Extensive experiments on GRAB and ARCTIC demonstrate consistent improvements in text adherence, motion quality, and physical plausibility over prior approaches, and ablations validate the contribution of each component.


\section*{Acknowledgements}
This work was supported by the National Research Foundation of Korea (NRF) grant funded by the Korean government (MSIT) (RS-2026-25499022), and Global -- Learning \& Academic research institution for Master’s·PhD students, and Postdocs (G-LAMP) Program of the NRF funded by the Ministry of Education (RS-2025-25442252).

%
%
\bibliographystyle{splncs04}
\bibliography{main}

\end{document}